\documentclass[10pt,twocolumn,letterpaper]{article}

\usepackage[pagenumbers]{cvpr} 

\definecolor{cvprblue}{rgb}{0.21,0.49,0.74}
\usepackage[pagebackref,breaklinks,colorlinks,allcolors=cvprblue]{hyperref}
\usepackage{listings}
\usepackage[T1]{fontenc}
\title{Compound Expression Recognition via Large Vision-Language Models}

\author{Jun Yu \quad Xilong Lu\\
University of Science and Technology of China\\
Hefei, Anhui, China\\
{\tt\small harryjun@ustc.edu.cn}, {\tt\small luxilong@mail.ustc.edu.cn}
}

\begin{document}
\maketitle
\begin{abstract}
Compound Expression Recognition (CER) is crucial for understanding human emotions and improving human-computer interaction. However, CER faces challenges due to the complexity of facial expressions and the difficulty of capturing subtle emotional cues. To address these issues, we propose a novel approach leveraging Large Vision-Language Models (LVLMs). Our method employs a two-stage fine-tuning process: first, pre-trained LVLMs are fine-tuned on basic facial expressions to establish foundational patterns; second, the model is further optimized on a compound-expression dataset to refine visual-language feature interactions. Our approach achieves advanced accuracy on the RAF-DB dataset and demonstrates strong zero-shot generalization on the C-EXPR-DB dataset, showcasing its potential for real-world applications in emotion analysis and human-computer interaction.
\end{abstract}    
\section{Introduction}
\label{sec:intro}

In the rapidly evolving field of artificial intelligence, emotion recognition has emerged as a critical research area, serving as a cornerstone for enabling natural human-computer interaction and deepening our understanding of human psychology \cite{Pantic2000}. Among its subfields, Compound Expression Recognition (CER), which focuses on identifying multiple co-occurring facial emotions, occupies a unique position. Human facial expressions rarely convey a single emotion; instead, they often reflect a blend of emotions, a complexity that CER aims to capture. This pursuit is not only academically intriguing but also holds significant practical relevance.

The applications of CER span a wide range of domains. In intelligent security systems, for instance, CER can play a transformative role. Consider a crowded airport where security cameras are ubiquitous. Advanced CER algorithms could detect a combination of anxiety and furtiveness in an individual's facial expression, potentially flagging a security threat and enabling proactive intervention \cite{Zhao2007}. In mental healthcare, CER offers a powerful tool for patient assessment. A psychiatrist might observe a blend of sadness and resignation in a patient's expression, providing valuable insights into their mental state and aiding in more accurate diagnosis and treatment planning \cite{Donato1999}. However, the inherent complexity of compound expressions, characterized by subtle nuances and intricate combinations of facial muscle movements, poses significant challenges, making CER a focal point of research.

Recent advancements in Large Vision-Language Models (LVLMs) have revolutionized multimodal data processing. These models are designed to seamlessly integrate visual data, such as facial features, with language data, such as emotion-related textual descriptions \cite{Brown2020, Radford2019}. This unique capability bridges the gap between vision and language, opening new avenues for CER. By learning from both modalities simultaneously, LVLMs can uncover previously hidden emotional patterns. For example, an LVLM might correlate the visual feature of a tightened jaw (associated with anger or stress) with the textual description "feeling pressured" to more accurately classify a compound emotion \cite{Brown2020, Radford2019}.

In the context of the Affective Behavior Analysis in-the-Wild (ABAW) competitions \cite{Kollias2025,kolliasadvancements,kollias20247th,kollias20246th,kollias2024distribution,kollias2023abaw2,kollias2023multi,kollias2023abaw,kollias2022abaw,kollias2021analysing,kollias2021affect,kollias2021distribution,kollias2020analysing,kollias2019expression,kollias2019deep,kollias2019face,zafeiriou2017aff}, significant progress has been made in CER and related tasks. The ABAW competitions have consistently pushed the boundaries of emotion recognition by introducing large-scale datasets and challenging tasks such as multi-label compound expression recognition, valence-arousal estimation, and action unit detection \cite{kollias2023multi, kollias2022abaw, kollias2021analysing}. These competitions have highlighted the importance of multi-task learning and the integration of diverse modalities for robust emotion recognition. For instance, the 7th ABAW competition focused on multi-task learning and compound expression recognition, emphasizing the need for models that can simultaneously handle multiple emotion-related tasks \cite{kollias20247th}. Similarly, the 6th ABAW competition demonstrated the effectiveness of distribution matching techniques for multi-task learning, which can be applied to CER to improve generalization across different emotional categories \cite{kollias20246th}.

In this paper, we propose a novel stage-wise LoRA (Low-Rank Adaptation) fine-tuning approach to optimize pre-trained models for CER, particularly in resource-constrained settings. LoRA, a technique that reduces the number of trainable parameters by introducing low-rank matrices, addresses the computational challenges of fine-tuning large models \cite{Hu2021}. Our stage-wise strategy enables the model to learn task-specific patterns more effectively. By adjusting the model's parameters in a phased manner, tailored to the data characteristics and task requirements at each stage, the model can better capture the intricate patterns of compound expressions, thereby enhancing its generalization capabilities.

Furthermore, we highlight the importance of well-designed context prompts for LVLMs in accurately performing CER. These prompts provide background knowledge and semantic guidance, helping the model focus on key features during recognition \cite{Devlin2019}. For instance, a context prompt might emphasize cultural variations in the interpretation of facial expressions, enabling the model to make more culturally sensitive and accurate predictions. This approach aligns with the findings from the ABAW competitions, where context-aware models have shown superior performance in recognizing complex emotional states \cite{kollias2021affect, kollias2023abaw}.

In summary, this paper leverages the strengths of LVLMs, a stage-wise LoRA fine-tuning strategy, and innovative context prompt design to develop an efficient and highly accurate CER method. Our goal is to address the longstanding challenges in this field and advance the practical application of emotion recognition technology, paving the way for more intelligent and empathetic human-machine interactions. The insights gained from the ABAW competitions, particularly in multi-task learning and context-aware modeling, provide a solid foundation for our approach \cite{kollias20246th, kollias2023multi, kollias2021affect}.
\section{Related Work}
\label{sec:Related}

In this section, we review existing research in Compound Expression Recognition (CER) and fine-tuning methods for Large Vision-Language Models (LVLMs). This review provides a comprehensive understanding of the current state of research and positions our proposed approach within the broader context.

\subsection{Compound Expression Recognition}
Compound Expression Recognition focuses on identifying multiple emotional states simultaneously expressed on the face. This research direction is of significant importance in artificial intelligence, as it captures the complexity of human emotions more accurately than single-emotion recognition. However, the high complexity and diversity of compound expressions, characterized by subtle variations and intricate combinations of facial muscle movements, pose substantial challenges for accurate recognition.

Early approaches to CER relied on handcrafted features. For example, geometric features of facial landmarks, such as the distance between eyebrows, the curvature of the mouth corners, and the degree of eye opening, were extracted and fed into traditional classifiers like Support Vector Machines (SVMs) for emotion classification \cite{Pantic2000, Bartlett1999}. However, these handcrafted features often fail to capture the nuanced and complex information inherent in compound expressions.

With the advent of deep learning, Convolutional Neural Networks (CNNs) have become widely adopted in CER. CNNs can automatically learn hierarchical features from facial images, significantly improving performance over traditional methods. Pre-trained architectures such as VGG-16 and ResNet have been directly applied to CER tasks \cite{Zhang2010}. Despite their success, these models still struggle with the high complexity of compound expressions. To address this, more advanced architectures have been proposed. For instance, multi-stream CNN models process different facial regions separately to capture local-global relationships in expressions \cite{Zhao2007}.

Recurrent Neural Networks (RNNs) and their variants, such as Long Short-Term Memory networks (LSTMs), have also been employed in CER. Given the temporal nature of facial expressions, RNN-based models can analyze dynamic changes over time, making them particularly effective for video-based CER tasks. These models infer potential emotions by analyzing sequences of facial changes, providing a richer understanding of compound expressions \cite{Donato1999}.

\subsection{Large Vision-Language Models (LVLMs)}
Large Vision-Language Models (LVLMs) have recently emerged as a transformative force in artificial intelligence, excelling in multimodal tasks by effectively integrating visual information (e.g., facial features) and language information (e.g., emotion-related text). LVLMs leverage large-scale data to build a deep understanding of both modalities, offering new opportunities for addressing complex tasks like CER.

Fine-tuning is a common strategy for adapting LVLMs to specific tasks. The most straightforward approach is full fine-tuning, where all parameters of the pre-trained LVLM are adjusted during training on the target task \cite{Devlin2019}. While this method can achieve high performance, it is computationally expensive and resource-intensive, particularly for large-scale LVLMs.

To mitigate these challenges, partial fine-tuning methods have been developed. One common strategy is to freeze the early layers of the LVLM and fine-tune only the later layers, as early layers typically capture general features while later layers are more task-specific \cite{Howard2018}. This approach significantly reduces the number of trainable parameters and computational costs.

Another popular method is adapter-based fine-tuning, which introduces small adapter layers into the pre-trained model. These adapters are trained independently while keeping the original model parameters fixed \cite{Houlsby2019}. This not only reduces computational overhead but also facilitates task switching, as different adapters can be trained for different tasks on the same base model.

In the realm of low-rank adaptation, the LoRA (Low-Rank Adaptation) method has gained significant attention. Unlike full fine-tuning, LoRA updates only a small number of low-rank matrices, drastically reducing memory usage and training time \cite{Hu2021}. LoRA has demonstrated strong performance in various fields, including natural language processing and vision-language tasks. By approximating weight updates through low-rank matrices, LoRA enables efficient fine-tuning of LVLMs for specific tasks like CER, making it particularly suitable for resource-constrained settings.
\section{Method}
Our approach for Compound Expression Recognition (CER) leverages the multimodal capabilities of Large Vision-Language Models (LVLMs) through a stage-wise LoRA (Low-Rank Adaptation) fine-tuning strategy and carefully designed context prompts. This methodology adapts LVLMs to the complex task of CER, enabling efficient and accurate recognition of compound facial expressions.

\begin{figure*}[h]
    \centering
    \includegraphics[width=1.0\textwidth]{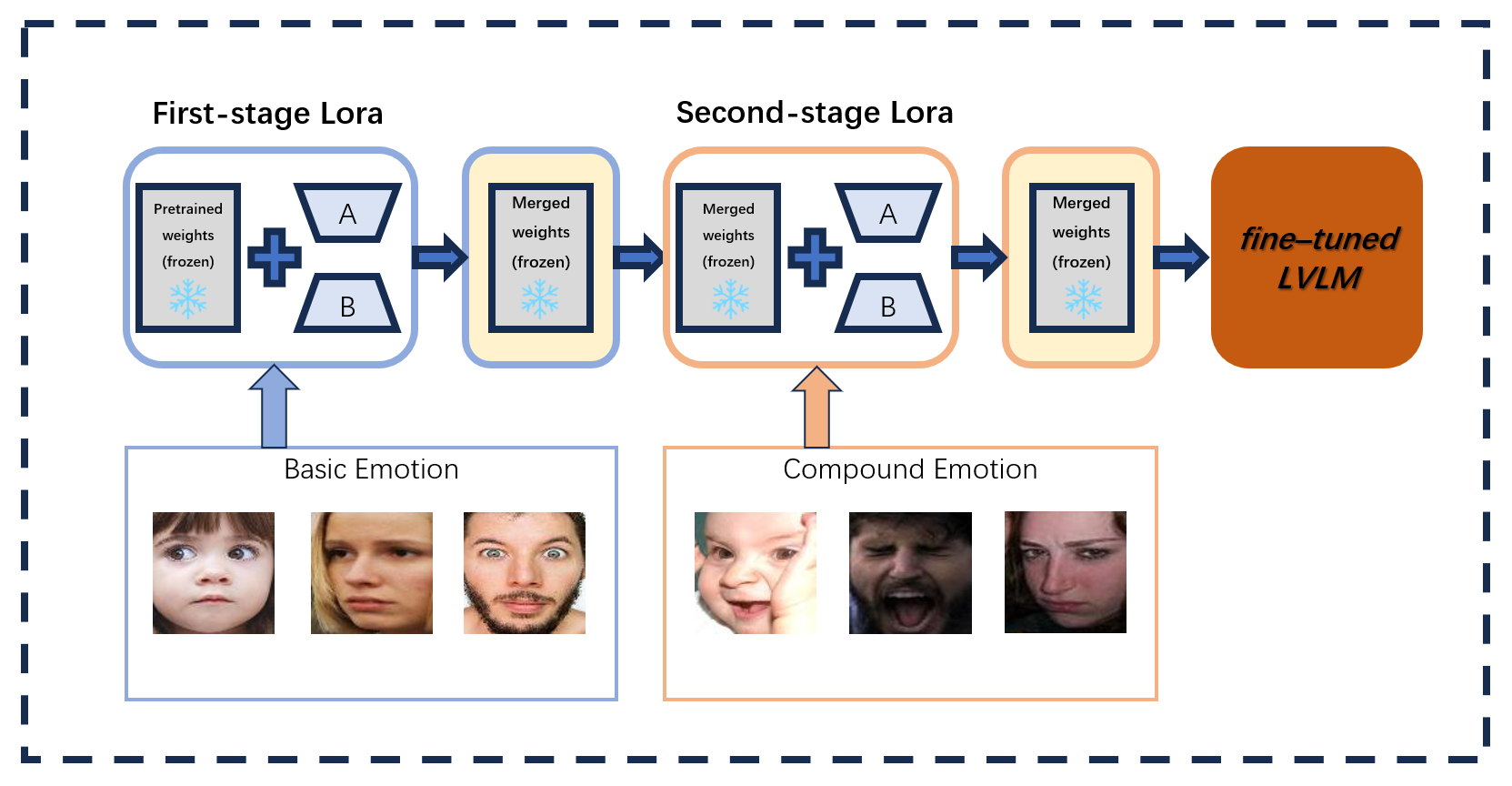}
    \caption{Stage-wise LoRA Fine-tuning Framework for Compound Expression Recognition}
    \label{fig:stage_lora_framework}
\end{figure*}

\subsection{Stage-wise LoRA Fine-Tuning}
\subsubsection{Low-Rank Adaptation (LoRA)}
LoRA is a parameter-efficient fine-tuning technique designed for large pre-trained models like LVLMs. In a standard neural network layer, the weight matrix \(W\) is high-dimensional, and updating all its parameters during fine-tuning can be computationally expensive and prone to overfitting. LoRA addresses this by introducing two low-rank matrices, \(A\) and \(B\), to approximate the weight update \(\Delta W\) as:
\begin{equation}
\Delta W \approx BA
\end{equation}
Here, \(A\) reduces the input dimensionality from \(d\) to \(r\) (where \(r \ll d\)), and \(B\) maps it back to the original dimension. The output \(y\) of the LoRA-modified layer is computed as:
\begin{equation}
y = W_0x + (BA)x
\end{equation}
where \(W_0\) is the original weight matrix. The number of trainable parameters is reduced from \(d^2\) to \(2dr\), significantly lowering computational costs. This reduction is formalized as:
\begin{equation}
N_{original} = d \times d = d^2
\end{equation}
\begin{equation}
N_{LoRA} = d \times r + r \times d = 2dr
\end{equation}
This efficiency makes LoRA particularly suitable for fine-tuning large-scale LVLMs in resource-constrained settings.

\subsubsection{First-stage Fine-Tuning on Basic Emotions}
In the first stage, we fine-tune the LVLM using the Basic subset of FER datasets. We initialize LoRA layers within the pre-trained LVLM and freeze the majority of its original parameters, updating only the low-rank matrices \(A\) and \(B\). This stage focuses on learning fundamental patterns of basic emotions, such as happiness, sadness, and anger, without overfitting to the idiosyncrasies of the pre-trained model. By capturing these foundational features, the model establishes a robust basis for recognizing more complex compound expressions in the subsequent stage.

\subsubsection{Second-stage Fine-Tuning on Compound Emotions}
In the second stage, we fine-tune the model using the Compound subset of FER datasets. We continue to update only the LoRA layers while keeping the original LVLM parameters frozen. This stage focuses on learning the intricate combinations of emotions present in compound expressions, such as "happily surprised" or "sadly angry." The two-stage approach enables the model to incrementally adapt to the increasing complexity of CER, building on its knowledge of basic emotions to accurately recognize compound expressions.

\subsection{Context Prompt Design}
\subsubsection{Design Principles}
Context prompts are designed to provide the LVLM with task-relevant information, guiding its attention to key facial features and their emotional significance. Our prompt design is guided by two key principles:
\begin{itemize}
    \item \textbf{Relevance}: Prompts should include descriptions directly related to facial expressions, such as how specific muscle movements correspond to emotions. For example, "A furrowed brow combined with a slightly open mouth may indicate a blend of anger and surprise."
    \item \textbf{Clarity}: Prompts should use clear and unambiguous language to ensure the model focuses on essential features for CER. This includes providing explicit instructions and examples to reduce ambiguity.
\end{itemize}

\subsubsection{Prompt Construction}
We construct context prompts by integrating natural-language descriptions of facial expressions and their associated emotions. Each prompt is structured to include:
\begin{itemize}
    \item \textbf{Task Objective}: A clear statement of the task, such as "Your task is to analyze the facial expression of the person(s) in the provided image and classify it into one of the seven predefined categories."
    \item \textbf{Category Definitions}: Descriptions of each compound emotion category, including key facial features. For example:
    \begin{itemize}
        \item \textit{Fearfully Surprised}: A mix of fear and surprise, characterized by widened eyes, raised eyebrows, and a slightly open mouth.
        \item \textit{Happily Surprised}: A blend of happiness and surprise, featuring a bright smile, raised eyebrows, and wide-open eyes.
        \item \textit{Sadly Surprised}: A combination of sadness and surprise, with downturned lips, raised eyebrows, and a look of shock.
    \end{itemize}
    \item \textbf{Analysis Guidelines}: Instructions for analyzing facial features, such as "Carefully examine the image to identify visible facial features like the eyes, eyebrows, mouth, and overall facial tension."
    \item \textbf{Output Format}: A template for the model's response, including both analysis and conclusion. For example:
    \begin{itemize}
        \item If no person is present: "Analysis: [Provide a detailed analysis of the image, noting the absence of any person.] Conclusion: There is no one in the image."
        \item If a person is present: "Analysis: [Provide a detailed analysis of the facial expression, describing the features that led to your conclusion.] Conclusion: The facial expression of the person in the image is '[Selected Category]'."
    \end{itemize}
\end{itemize}

\subsection{Incorporation into the Inference Process}
During inference, we append the context prompt to each test image input. The LVLM processes this combined input (image + prompt) to predict the compound emotion. The prompt serves as a semantic guide, directing the model's attention to relevant facial features and ensuring a structured output format. For example, given an image of a person with wide-open eyes, raised eyebrows, and a bright smile, the model might generate the following output:

\begin{lstlisting}
Analysis: The person in the image has wide-open eyes, raised eyebrows, and a bright smile, 
indicating a mix of happiness and surprise.
Conclusion: The facial expression of the person in the image is 'Happily Surprised'.
\end{lstlisting}

This structured approach enhances the model's ability to accurately recognize compound expressions and provide interpretable results.
\section{Experiments}
\subsection{Datasets}
\subsubsection{RAF-DB}
We utilize the Basic and Compound subsets from the RAF-DB (Real-world Affective Faces Database). The Basic subset within RAF-DB contains a substantial number of facial images labeled with basic emotions, including happiness, sadness, anger, surprise, fear, and disgust. These serve as a fundamental resource for the model to learn basic facial expression patterns. The Compound subset, on the other hand, is composed of images with more complex, compound emotional expressions. Each image in this subset is precisely annotated with the specific combination of emotions present, equipping the model to learn the subtleties of compound expressions.

\subsubsection{Aff-Wild2}
Aff-Wild2 is a large-scale in-the-wild dataset designed for affective computing tasks, including emotion recognition and valence-arousal estimation. It contains over 1,000 videos with more than 2.5 million frames, annotated for 7 basic emotions (anger, disgust, fear, happiness, sadness, surprise, and neutral) and 6 compound emotions. The dataset is highly diverse, covering a wide range of ages, ethnicities, and lighting conditions, making it suitable for training robust models for real-world applications. We use Aff-Wild2 in conjunction with RAF-DB for training, leveraging its large-scale and diverse annotations to enhance the model's generalization ability.

\subsubsection{C-EXPR-DB}
C-EXPR-DB stands as the largest and most diverse in-the-wild audiovisual database to date. It encompasses 400 videos, totaling around 200,000 frames, meticulously annotated for 12 compound expressions and various affective states. Additionally, C-EXPR-DB includes annotations for continuous valence-arousal dimensions [-1, 1], speech detection, facial landmarks and bounding boxes, 17 action units (facial muscle activation), and facial attributes. In the Compound Expression (CE) Recognition Challenge, a total of 56 unlabeled videos were selected, covering 7 types of compound expressions. The extracted video tags consist of seven compound expressions: Fearfully Surprised, Happily Surprised, Sadly Surprised, Disgustedly Surprised, Angrily Surprised, Sadly Fearful, and Sadly Angry. C-EXPR-DB serves as the final test set for evaluating the model's performance in a real-world competition setting.

\subsection{Implement Details}
\subsubsection{Model Selection}
We selected Qwen-vl as the base model for our experiments. This model has demonstrated strong performance in various multimodal tasks, making it a suitable choice for compound expression recognition. We initialized the model with pre-trained weights on large-scale vision language datasets, which provides the model with a good starting point for learning facial expression-related features.

\subsubsection{LoRA Configuration}
In the stage-wise LoRA fine-tuning process, we configured the LoRA layers. For the first-stage fine-tuning on the Basic subset, we set the rank of the low-rank matrices to 16. This value was chosen based on preliminary experiments to balance the reduction of trainable parameters and the model's representational ability. The learning rate was set to 0.0001. During the second-stage fine-tuning on the Compound subset, the rank of the low-rank matrices was adjusted to 8, and the learning rate was set to 0.0001. These adjustments were made to better adapt the model to the increasing complexity of compound expressions.

\subsubsection{Context Prompt Generation}
The context prompts were generated using a rule-based approach. We defined a set of templates related to facial expression analysis. For example, for an image, the prompt might start with "Analyze the following facial image to identify compound emotions. Look for signs of key facial features related to basic emotions and determine how they combine." The generation process was automated, and the prompts were carefully designed to provide relevant information to the model without over-specifying the answer.

\subsubsection{Training and Validation}
The training was conducted on RTX3090 GPUs to accelerate the process. The batch size was set to 1 for both the first-stage and second-stage fine-tuning. The number of epochs for the first-stage fine-tuning was 20, and for the second-stage fine-tuning was 10.

We train our model on a combination of the \textbf{Aff-Wild2} and \textbf{RAF-DB} datasets, leveraging the large-scale and diverse annotations from Aff-Wild2 to enhance the model's generalization ability. The model is then validated on the \textbf{RAF-DB} dataset to ensure its performance on both basic and compound emotion recognition tasks. Finally, the model is evaluated on the \textbf{C-EXPR-DB} dataset, which serves as the final test set for the Compound Expression Recognition Challenge. This approach ensures that the model is robust and capable of handling real-world variability in facial expression recognition.
\section{Results}
\subsection{RAF-DB Basic Emotion Recognition}

To adapt to the real-world image environment of the C-EXPR-DB dataset, we conducted experiments on both \textbf{aligned images} and \textbf{original images} for RAF-DB Basic Emotion Recognition. The results are presented in Table~\ref{tab:basic_emotion}.

\begin{table}[ht]
\centering
\caption{Accuracy for RAF-DB Basic Emotion Recognition}
\label{tab:basic_emotion}
\begin{tabular}{|l|c|c|}
\hline
\textbf{Emotion} & \textbf{Aligned (\%)} & \textbf{Original (\%)} \\ \hline
Sadness          & 88.12                & 86.89                 \\ \hline
Surprise         & 94.45                & 92.58                 \\ \hline
Happiness        & 96.50                & 94.39                 \\ \hline
Disgust          & 68.25                & 64.88                 \\ \hline
Anger            & 83.30                & 80.78                 \\ \hline
Fear             & 73.60                & 69.22                 \\ \hline
Neutral          & 92.20                & 90.65                 \\ \hline
\textbf{Overall} & \textbf{89.78}       & \textbf{87.84}        \\ \hline
\end{tabular}
\end{table}

As shown in Table~\ref{tab:basic_emotion}, the model achieves higher accuracy on aligned images compared to original images across all emotion categories. For example, the accuracy for \textbf{Happiness} increases from 94.39\% on original images to 96.50\% on aligned images. Similarly, the overall accuracy improves from 87.84\% to 89.78\%. This indicates that aligned images, with their standardized facial features, provide a more consistent and reliable input for emotion recognition. However, the performance on original images remains strong, demonstrating the model's robustness in real-world scenarios.

\subsection{RAF-DB Compound Emotion Recognition}

For RAF-DB Compound Emotion Recognition, we also conducted experiments on both \textbf{aligned images} and \textbf{original images}. The results are presented in Table~\ref{tab:compound_emotion}.

\begin{table}[ht]
\centering
\caption{Accuracy for RAF-DB Compound Emotion Recognition}
\label{tab:compound_emotion}
\begin{tabular}{|l|c|c|}
\hline
\textbf{Emotion}           & \textbf{Aligned (\%)} & \textbf{Original (\%)} \\ \hline
Happily Surprised          & 90.25                & 88.52                 \\ \hline
Sadly Disgusted            & 75.30                & 71.70                 \\ \hline
Happily Disgusted          & 80.45                & 78.09                 \\ \hline
Fearfully Angry            & 77.80                & 73.64                 \\ \hline
Angrily Disgusted          & 94.10                & 92.18                 \\ \hline
Angrily Surprised          & 50.20                & 41.58                 \\ \hline
Sadly Surprised            & 52.40                & 43.33                 \\ \hline
Fearfully Surprised        & 80.60                & 78.10                 \\ \hline
Disgustedly Surprised      & 55.80                & 45.71                 \\ \hline
Sadly Fearful              & 60.30                & 55.45                 \\ \hline
Sadly Angry                & 55.20                & 46.36                 \\ \hline
\textbf{Overall}           & \textbf{78.50}       & \textbf{74.39}        \\ \hline
\end{tabular}
\end{table}

As shown in Table~\ref{tab:compound_emotion}, the model achieves higher accuracy on aligned images compared to original images for all compound emotion categories. For example, the accuracy for \textbf{Angrily Disgusted} increases from 92.18\% on original images to 94.10\% on aligned images. Similarly, the overall accuracy improves from 74.39\% to 78.50\%. This suggests that aligned images provide a more consistent representation of compound emotions, which are inherently more complex and challenging to recognize. However, the performance on original images remains competitive, highlighting the model's ability to handle real-world variability.

\section{Conclusion}

The results demonstrate that aligned images consistently outperform original images in both basic and compound emotion recognition tasks. This is expected, as aligned images provide standardized facial features, reducing variability and improving model performance. However, the strong performance on original images underscores the model's robustness and generalizability to real-world scenarios. Future work could focus on further bridging the gap between aligned and original image performance, potentially through domain adaptation techniques or more sophisticated data augmentation strategies.
{
    \small
    \bibliographystyle{ieeenat_fullname}
    \bibliography{main}
}


\end{document}